\pdfoutput=1
\documentclass[11pt]{article}
\usepackage{graphicx}
\usepackage{float}
\usepackage[preprint]{acl}

\usepackage{times}
\usepackage{latexsym}

\usepackage[T1]{fontenc}
\usepackage[utf8]{inputenc}

\usepackage{microtype}

\usepackage{inconsolata}

\usepackage{graphicx}
\usepackage{url}
\usepackage{algorithm}
\usepackage{algorithmic}
\usepackage{enumitem}
\usepackage{tikz}
\definecolor{light_gray}{HTML}{FFFFFF} % F5F5F5
\definecolor{solid_gray}{HTML}{000000} % previous color: d4d4d4
\usepackage{multirow}
\usepackage{array}
\usepackage{hhline}
\usepackage{pifont}
\usepackage{makecell}
\usepackage{array} % for vertical cell alignment
\usepackage{booktabs}
\usepackage{siunitx}
\usepackage{nameref}
\usepackage{xcolor}
\usepackage{transparent}
\usepackage{soul}
\usetikzlibrary{calc}
\usepackage{tcolorbox}
\usepackage{amssymb}
\usepackage{bbding}
\usepackage{pifont}
\usepackage{longtable}
\usepackage{tabularx}
\usepackage{rotating}
\usepackage{adjustbox}
\usepackage{bibentry}

\usepackage{amsmath}

\usepackage{comment}
\usepackage{float} 
\usetikzlibrary{positioning}

\usepackage{tikz}
\usetikzlibrary{shapes}

\newtcbox{\highlight}[1][]{%
    colback=green!15!white,
    colframe=green!15!white,
    boxrule=0pt,
    boxsep=0pt,
    left=2pt,
    right=2pt,
    top=2pt,
    bottom=2pt,
    sharp corners,
    #1
}
\title{In-Context Learning for Long-Context Sentiment Analysis on Infrastructure Project Opinions}

\author{First Author \\
  Affiliation / Address line 1 \\
  Affiliation / Address line 2 \\
  Affiliation / Address line 3 \\
  \texttt{email@domain} \\\And
  Second Author \\
  Affiliation / Address line 1 \\
  Affiliation / Address line 2 \\
  Affiliation / Address line 3 \\
  \texttt{email@domain} \\}

\author{
  \textbf{Alireza Shamshiri\textsuperscript},
  \textbf{Kyeong Rok Ryu\textsuperscript},
  \textbf{June Young Park\textsuperscript} \\
  The University of Texas at Arlington, Arlington, TX, USA \\
  \small{
    \texttt{\{alireza.shamshiri, kyeongrok.ryu, juneyoung.park\}@uta.edu}
  }
}

\begin{document}
\maketitle
\begin{abstract}
Large language models (LLMs) have achieved impressive results across various tasks. However, they still struggle with long-context documents. This study evaluates the performance of three leading LLMs: GPT-4o, Claude 3.5 Sonnet, and Gemini 1.5 Pro on lengthy, complex, and opinion-varying documents concerning infrastructure projects, under both zero-shot and few-shot scenarios. Our results indicate that GPT-4o excels in zero-shot scenarios for simpler, shorter documents, while Claude 3.5 Sonnet surpasses GPT-4o in handling more complex, sentiment-fluctuating opinions. In few-shot scenarios, Claude 3.5 Sonnet outperforms overall, while GPT-4o shows greater stability as the number of demonstrations increases.
\end{abstract}

\section{Introduction}

Large language models (LLMs) have demonstrated human-level performance across various tasks, such as coding, question answering, mathematical problem-solving, classification, and sentiment analysis (SA) \cite{valmeekam2022large, shi2022language, navard2024knobgen, chen2024humans, bavaresco2024llms, wankhade2022survey}. Despite the emergence of LLMs and their in-context learning (ICL) capabilities, several critical challenges persist in leveraging LLMs for domain-specific SA task such as processing long input sequences, managing frequent sentiment shifts, and accommodating domain-specific terminology across diverse contexts \cite{medhat2014sentiment, wankhade2022survey, raghunathan2023challenges, zhu2024model, simmering2023large, zhang2023sentiment}.

Furthermore, to fully understand the true performance and capabilities of LLMs in various tasks, particularly SA, it is crucial to consider data contamination as an important factor which is generally disregarded. Data contamination refers to the inclusion of test data in the training dataset, which can artificially inflate the performance of these models in zero-shot and few-shot scenarios \citep{golchin2023time, deng2023investigating, golchin2023data, dong2024generalization, xu2024benchmark, golchin2024memorization}.

In response, we study the performance of three leading LLMs on long-context, complex, and sentiment-fluctuating documents across both potentially contaminated and uncontaminated datasets in the domain of public infrastructure projects. To achieve this, we perform experiments in zero-shot and few-shot settings, including three-, six-, and nine-shot scenarios.

The key contributions of this paper include:

\textbf{(1)} We evaluate the performance of three prominent LLMs in handling the SA task within the domain of infrastructure projects, focusing on long-context and complex documents that exhibit frequent sentiment fluctuations. This assessment provides a rigorous test of the models' true performance in a domain-specific SA task, and highlighting the domain-specific challenges in using LLMs.

\textbf{(2)} We investigate the impact of data contamination on LLMs in zero-shot and few-shot settings to examine how prior exposure to data affects the models' performance in performing SA task when provided with relevant examples. This helps us genuinely assess the models' performance.

\section{Related Work}

\setlength{\arrayrulewidth}{0.3mm} % Keep this for standard lines
\begin{table*}[h!]
\centering
\fontsize{25}{35}\selectfont
\resizebox{\textwidth}{!}{ % Add this line to resize the table
\begin{tabular}{ccccccccccc}
\Xhline{1mm} % Top thicker line
\textbf{Dataset} &
  \textbf{\begin{tabular}[c]{@{}c@{}}No.\\  \end{tabular}} &
  \textbf{\begin{tabular}[c]{@{}c@{}}Mean \\ \end{tabular}} &
  \textbf{\begin{tabular}[c]{@{}c@{}}SD \\ \end{tabular}} &
  \textbf{\begin{tabular}[c]{@{}c@{}}Min.\\ \end{tabular}} &
  \textbf{\begin{tabular}[c]{@{}c@{}}25th Pctl. \\ \end{tabular}} &
  \textbf{\begin{tabular}[c]{@{}c@{}}Median \\ \end{tabular}} &
  \textbf{\begin{tabular}[c]{@{}c@{}}75th Pctl.\\ \end{tabular}} &
  \textbf{\begin{tabular}[c]{@{}c@{}}Max.\\ \end{tabular}} &
  \textbf{\begin{tabular}[c]{@{}c@{}}Contamination\\ \end{tabular}} &
  \textbf{\begin{tabular}[c]{@{}c@{}}Complexity Rkg.\\ \end{tabular}} \\ \hline
Facebook (FB)        & 230 & 436  & 602  & 15  & 153  & 274  & 489  & 5276  & \checkmark & 1 \\
News (NS)            & 230 & 3500 & 2248 & 70  & 1764 & 2977 & 5032 & 10538 & \checkmark & 3 \\
Public Hearing (PH)  & 230 & 660  & 1558 & 43  & 210  & 345  & 564  & 13652 & $\times$ & 2 \\
Scoping Meeting (SC) & 230 & 980  & 1741 & 12  & 215  & 464  & 1041 & 17921 & $\times$ & 4 \\ \hline
\Xhline{1mm} % Bottom thicker line
\end{tabular}
} % Close the resizebox
\caption{Length distribution and percentile range of datasets, contamination status (× = contamination-free, \checkmark = probable contamination), and complexity ranking (4 = most complex, 1 = least complex). The detailed assessments of complexity ranking and contamination status are described in Appendices C and D.}
\label{tab:dataset}
\end{table*}

ICL has emerged as a novel approach that enhances the performance of LLMs without requiring additional training or weight updates~\cite{brown2020language, openai2024gpt, dong2022survey}. However, despite its widespread use, the underlying mechanisms of ICL and its effectiveness in improving performance remain unclear~\cite{jiang2023latent, dai2022can}. ICL is generally performed in few-shot and many-shot learning regimes~\cite{agarwal2024many, li2023context, anil2024many}. This approach is applied to a range of tasks such as math problems, using LLMs as judges, reasoning, question answering, grading, classification, and more~\cite{valmeekam2022large, shi2022language, golchin2024grading, saparov2022language, chen2024humans, bavaresco2024llms}. Among various classification tasks, several studies have examined the performance of LLMs using ICL for SA~\citep{yang2024faima, zhan2024optimization, 
shaikh2023exploring}. 

The performance evaluation of the GPT-4 and Flan-T5 models on various types of sentiment classification tasks, conducted by \citet{zhang2023sentiment}, indicated that, although LLMs excelled at simple tasks and performed better than smaller models in few-shot learning, they still struggled with complex analyses~\cite{zhang2023sentiment}.

It was found that both GPT-3.5 and GPT-4 deliver the best performance in zero-shot cross-lingual SA~\cite{zhu2024model}. Additionally, GPT-4 and GPT-3.5 have undergone performance assessments for the aspect-based sentiment analysis (ABSA) task through zero-shot, few-shot, and fine-tuned settings \cite{simmering2023large}. The results demonstrated the great potential of fine-tuned LLMs for ABSA, with GPT-3.5 performing particularly well in fine-tuning tasks. In the zero-shot condition, GPT-4 showed slightly lower performance compared to GPT-3.5 fine-tuned. However, GPT-4's performance significantly improved when provided with in-context examples, surpassing GPT-3.5 in few-shot settings.

A large body of research has been developed to evaluate the performance of LLMs on various long-context tasks~\cite{liu2024lost}. Meanwhile, due to the limited effective context window size of LLMs during pretraining for handling long-context tasks, various approaches have been proposed to extend the context window of LLMs, such as token analysis and efficient fine-tuning optimization strategies~\cite{hu2024longrecipe, chen2023longlora, liu2024lost, chen2023extending, peng2023does}.

On the other hand, although the performance of LLMs on various SA tasks has been investigated, it remains unclear to what extent they can effectively perform SA on long, complex, and opinion-varying documents, particularly in domain-specific contexts such as infrastructure projects~\cite{jiang2016public, zeng2023public, da2021proposal, kim2021public, baek2023automated}. Therefore, we investigate the performance of three LLMs to study the aforementioned gaps.

\section{Approach}

In this study, we scraped and collected public opinions on the North Houston Highway Improvement Project from four sources due to its high controversy and data availability. This included scraping 230 instances each from Facebook (FB) posts and news articles (NS), as well as gathering field data and opinions from scoping meeting (SC) comments, and public hearing (PH) comments using comment cards, surveys, emails, and other sources. 

As illustrated in Table~\ref{tab:dataset}, the FB dataset is the shortest and least complex, with a mean length of 436 words, while NS is the longest. PH documents fall in the middle, with a mean of 660 words, while SC average 980 words, making it the second longest dataset, with the most complex language and opinion-varying documents.

To provide a comprehensive evaluation, we evaluate three models including GPT-4o, Claude 3.5 Sonnet, and Gemini 1.5 Pro under two settings: zero-shot and few-shot, using three-, six-, and nine-shot prompts.

In the zero-shot setting, models rely solely on their pre-existing knowledge without any provided examples, as illustrated in Figure 3 in Appendix \ref{app:Data Labeling}. 

For the few-shot setting, three-shot prompting includes one example per class (one positive, one negative, and one neutral), as shown in Figure 4 in Appendix \ref{app:Data Labeling}. Six-shot prompting expands the three-shot setup by adding three more examples from each class, and nine-shot prompting maximizes context by adding three additional examples for each class. These examples are intended to enhance the model's sentiment classification accuracy. 

It is essential to emphasize that prompt instances are randomly selected from the associated dataset to mitigate potential user bias in example selection and to ensure consistency across all datasets. After the completion of the experiments, the selected instances are excluded from the entire dataset for performance measurements, resulting in a total of 221 instances per dataset.

\section{Results and Discussion}

This section presents the results and discussion of the zero-shot and few-shot findings.

\subsection{Zero-Shot Performance Evaluation}

\begin{figure}
    \centering
    \includegraphics[width=1\linewidth]{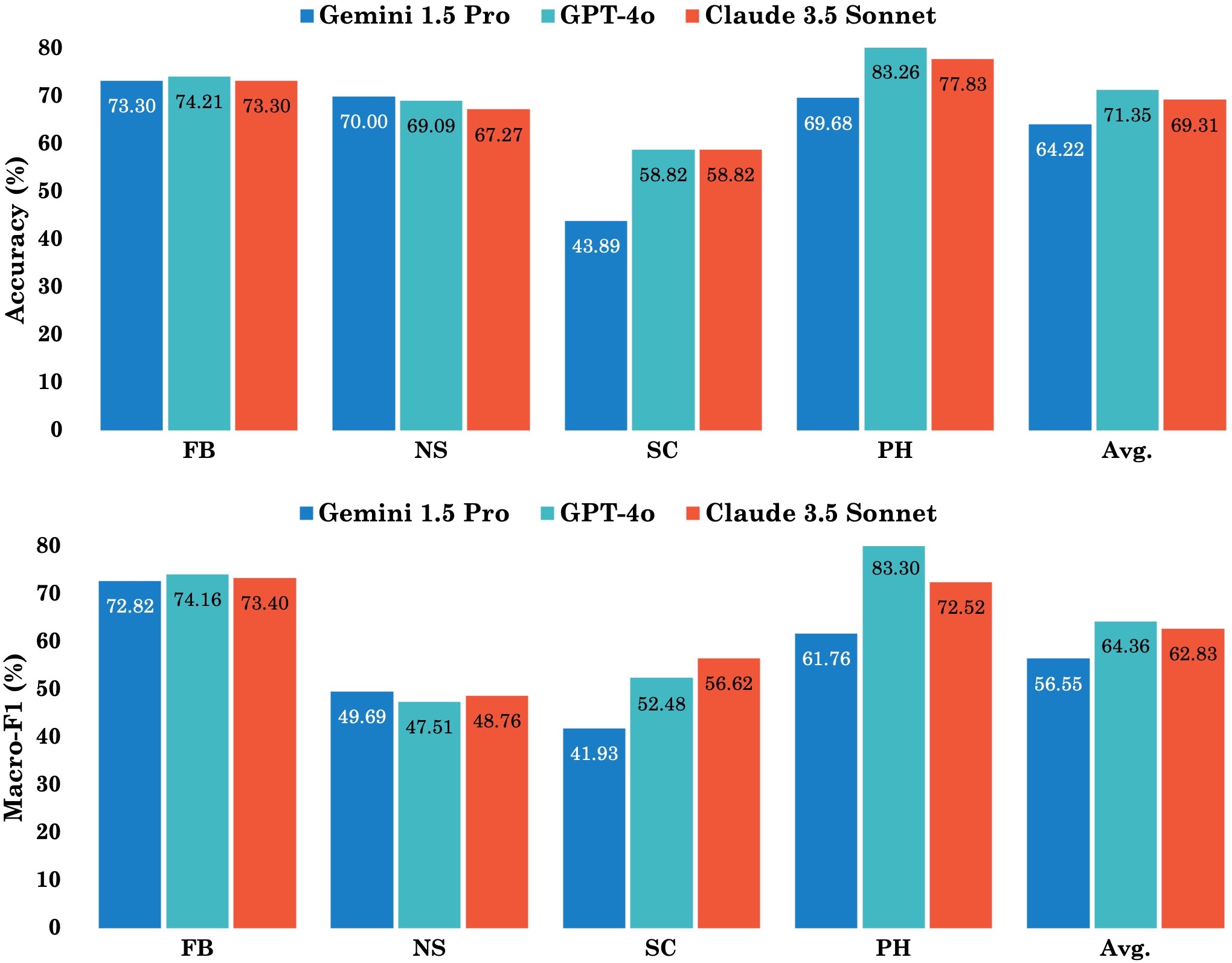}
    \caption{Performance comparison of zero-shot results. The average is calculated based on each model's performance across all four datasets}
    \label{Zero-Shot}
\end{figure}

On average, although GPT-4o outperforms both Claude 3.5 Sonnet and Gemini 1.5 Pro on simpler and shorter datasets, Claude 3.5 Sonnet achieves higher performance on more complex, sentiment-wavering, and lengthier datasets such as NS and SC as shown in Figure~\ref{Zero-Shot}. It can be seen that while Gemini performs satisfactorily on the longest dataset (NS), it achieves the least performance on the other datasets.

All models achieve relatively low and similar performance on the NS dataset, which has the longest context and simple language, with Gemini 1.5 Pro outperforming the others and Claude 3.5 Sonnet performing comparatively lower. On the other hand, GPT-4o achieved the lowest performance on the NS dataset, which might be attributed to the size of the model's parameters compared to the other two models. It can be said that, despite the contamination and simplicity of the language in the NS dataset, the models still struggle to analyze SA in long-context comments.

Despite the FB and PH datasets having almost similar mean comment lengths, the PH dataset comment lengths are much more spread out. Additionally, the PH dataset contains more complex and sentiment-wavering comments compared to FB. The differences in performance among the three models are relatively minor on the FB dataset. However, in the more complex and lengthier PH dataset, the performance differences across the models are notably higher. This variation might be attributed to the dataset's contamination, where the models lack prior knowledge, in contrast to the FB dataset results, or due to the more dynamic and ever-changing sentiments in the PH comments compared to FB. GPT-4o achieved the highest performance on both FB and PH datasets, while Claude 3.5 Sonnet performed relatively lower in zero-shot sentiment analysis classification.

In the SC dataset, which is free from data contamination and features the most challenging language with frequent shifts in opinions, all models achieved relatively low performance. However, Claude 3.5 Sonnet outperformed the other models, while Gemini 1.5 Pro recorded the lowest performance at 41.93\%.

\begin{figure*}[ht]
    \centering
    \includegraphics[width=\textwidth]{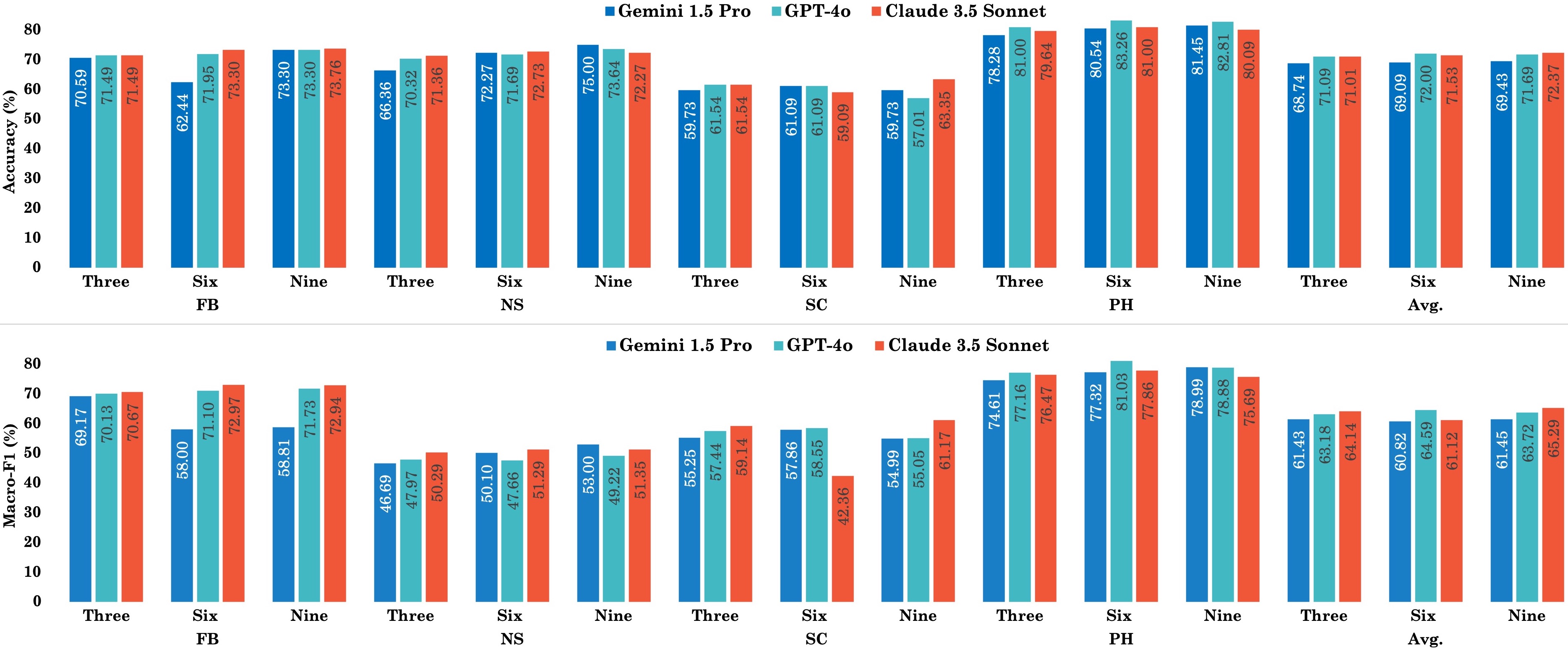}
    \caption{Performance comparison of few-shot results, with average outcomes calculated to provide deeper insights.}
    \label{few-shot}
\end{figure*}

\subsection{Few-Shot Performance Evaluation}

The few-shot analysis highlights that GPT-4o and Claude 3.5 Sonnet exhibit similar performance trends, and their results remain competitive, as shown in Figure~\ref{few-shot}. However, despite Claude 3.5 Sonnet achieving higher overall performance, particularly in the nine-shot setting for the majority of datasets, GPT-4o does not benefit from an increased number of demonstrations for improved performance, as there are only slight changes in performance across the three-shot, six-shot, and nine-shot settings. Nevertheless, performance mostly fluctuates and declines across all four datasets for all three models in the six-shot and nine-shot settings.

This fluctuation in performance with an increasing number of shots could be attributed to several factors. Although existing studies generally suggest that increasing the number of demonstration examples enhances in-context learning performance \cite{liu2021makes}, the results show that LLMs still struggle with SA task in long contexts containing multiple pieces of information, leading to inconsistent or degraded performance.

Furthermore, it can be inferred that as input prompts become longer, model performance may either decline or fluctuate, which highlights the challenges of LLMs due to the maximum sequence length encountered during training. This limitation hampers their ability to handle task requiring deep comprehension of complex and lengthy instructions.

Another reason for this phenomenon might be the challenges LLMs face in low-resource domain. This stems from their struggle to fully grasp specialized terminologies and nuances unique to specific domains, which results in inconsistent performance, particularly in interpreting highly domain-specific information. Without sufficient exposure to domain-specific examples during prompt demonstrations, the model may fail to generalize effectively.

Gemini 1.5 Pro demonstrates notable performance improvement in few-shot settings compared to GPT-4o and Claude 3.5 Sonnet in NS, similar to the observations in zero-shot results. A general trend shows that as the complexity and length of datasets decrease, the increase in model performance with more shots declines and fluctuates in SC, PH, and FB, respectively. This might indicate that the performance of the Gemini model is more affected by dataset length rather than complexity or the number of sentiment shifts for performance improvement.

Another observation, when comparing the results of zero-shot with few-shot (three-shot) settings, considering the data contamination status of the datasets, shows higher or similar performance on contaminated datasets compared to uncontaminated datasets, which achieved lower performance in zero-shot compared to three-shot. The contaminated dataset results likely reflect memorization, while the contamination-free results reflect true learning and adaptation. This distinction highlights the importance of few-shot learning for improving performance on uncontaminated and unfamiliar datasets.

\section{Conclusion}

The zero-shot analysis indicates that GPT-4o excels in simple, short SA task across various datasets, while Claude 3.5 Sonnet outperforms it in more complex, sentiment-wavering task. In the few-shot analysis, both models exhibit similar trends, with Claude 3.5 Sonnet achieving superior results on most datasets. However, GPT-4o demonstrates greater stability with an increasing number of shots, and Gemini 1.5 Pro performs well on longer datasets in both zero-shot and few-shot scenarios.

We found that the models exhibit comparable or superior performance in zero-shot settings on contaminated datasets compared to uncontaminated ones. While contaminated datasets can cause an increase in memorization and performance in zero-shot scenarios, the models' performance often declines or remains unchanged as the number of shots increases across four datasets which highlights the difficulties that LLMs experience due to the maximum sequence length, restricting their capability to comprehend  intricate and lengthy prompts.

\section{Limitations}

While this study aims to evaluate the performance of LLMs in analyzing complex, lengthy, and sentiment-wavering public opinions, several limitations remain. First, the scope of the evaluation is restricted by the limited number of shots, which may affect the robustness of the results. Expanding the number of shots and conducting many-shot analysis could provide more comprehensive insights. Second, this study does not explore a wider range of models, which could offer a broader comparison of performance across different architectures and versions. Finally, the evaluation is limited to overall sentiment classification task and does not extend to ABSA, which would provide a more granular understanding of opinion sentiment.

\bibliography{custom}

\appendix

\section{Experimental Setup}

For all experiments, we use {\fontfamily{qcr}\selectfont gpt-4o} endpoint for GPT-4o, {\fontfamily{qcr}\selectfont gemini-1.5-pro} endpoint for Gemini 1.5 Pro, and {\fontfamily{qcr}\selectfont claude-3-5-sonnet-20240620} endpoint for Claude 3.5 Sonnet. To promote deterministic outputs from the selected models, we set the temperature to 0 and the final results are the averages of three independent runs.

\section{Data Labeling}
\label{app:Data Labeling}
To ensure accurate sentiment labeling for the four datasets, we employed a majority voting approach~\cite{feldman2006majority}. Each dataset contains 230 instances, and for each instance, three experts annotate sentiment labels with three classes: positive, negative, or neutral. The final label for each instance is determined through a majority voting mechanism, wherein the most frequently assigned label is selected as the final label. We did not collect, use, or process any personal information at any stage. All data scraping, collection methods, and data labeling excluded personal identifiable information from the data. We performed this manually.

We calculated Fleiss' Kappa to assess the overall agreement among the three annotators~\cite{falotico2015fleiss}. The results are shown in Table 2, which includes the Fleiss' Kappa values along with the corresponding agreement levels.

\begin{table}[h]
\centering
\small
\begin{tabular}{ccc}
\toprule
\textbf{Dataset} & \textbf{Fleiss' K} & \textbf{Agreement Level} \\ 
\midrule
\textbf{FB} & 0.70 & Substantial Agreement \\
\textbf{NS} & 0.62 & Substantial Agreement \\
\textbf{SC} & 0.55 & Moderate Agreement \\
\textbf{PH} & 0.67 & Substantial Agreement \\ 
\bottomrule
\end{tabular}
\caption{Fleiss' Kappa results for agreement on labeled sentiments for each dataset}
\label{labelss}
\end{table}

Figure~\ref{figure:ZCoT-with-answers1} shows a comment randomly selected to be used for zero-shot setting and Figure~\ref{figure:ZCoT-with-answers2} illustrate the randomly selected instances and associated final labels for three-shot settings.
%\vspace{-10pt}

\begin{figure}[H]
    \centering
    \begin{minipage}{0.48\textwidth}
        \centering
        \begin{tikzpicture}[rounded corners=8pt, thick, text=black, text opacity=1]
            \node[draw=solid_gray, fill=light_gray, line width=1pt, 
            text=black, text width=0.92\textwidth, align=left, 
            font=\fontsize{6.5pt}{9pt}\selectfont, inner xsep=5pt, 
            inner ysep=5pt] at (0,0) {\textbf{Instruction:}
            % Sample input prompt.
            As a sentiment analyzer, determine the overall sentiment expressed in the following comments. Your response should be one of the following values: positive, neutral, or negative. Provide only the sentiment value in lowercase.

            - - -

            \textbf{Comment:}

            We support the project to improve traffic flow and reduce congestion in Houston. This project will enhance the corridor for economic development and benefit the employees of our firm. Furthermore, we strongly recommend the use of Design-Build contracting to accelerate construction and offer a variety of design and construction solutions for TxDOT. Lastly, the project will create much-needed jobs for our industry, which will provide a local economic boost.

            - - -
            
            \textbf{Sentiment: }
            
            \highlight{Positive}
            };
        \end{tikzpicture}
        \caption{An illustration of the zero-shot prompt along with an answer generated by a LLM model.}
        \label{figure:ZCoT-with-answers1}
\end{minipage}
\end{figure}

\begin{figure}
\begin{minipage}{0.48\textwidth}
    \centering
        \begin{tikzpicture}[rounded corners=8pt, thick, text=black, text opacity=1]
                    \node[draw=solid_gray, fill=light_gray, line width=1pt, 
                    text=black, text width=0.92\textwidth, align=left, 
                    font=\fontsize{6.5pt}{9pt}\selectfont, inner xsep=5pt, 
                    inner ysep=5pt] at (0,0) {\textbf{Instruction:}
                    % Sample input prompt.
                    As a sentiment analyzer, determine the overall sentiment expressed in the following comments. Your response should be one of the following values: positive, neutral, or negative. Provide only the sentiment value in lowercase. Here are examples of true sentiments for comments that you should use to refine your analysis:
        
                    - - -
        
                    \textbf{Comment Instances and Associated Sentiment Labels:}
        
                    \definecolor{lightcyan}{rgb}{0.88, 1.0, 1.0}
        
                    % Set the highlight color to light cyan
                    \sethlcolor{lightcyan}
        
                    \noindent\textbf{Comment:} 
        
                    I'm highly against rerouting I-45 around the east side of downtown. The current Pierce has congestion partly because of the hard curves in the existing highway, causing traffic to brake. The 'replacement' would have even more awkward curves that would introduce more congestion without fixing the problems, as well as uneconomical ROW acquisitions.
        
                    \noindent\textbf{Sentiment:} Negative
        
                    \noindent\textbf{Comment:} 
        
                    I support the North Houston Highway Improvement Project. I would like to thank TxDOT and its consultant for the substantial improvements in the design since the original design was presented in 2015. 3. Due to the high cost and construction disruption of this project, TxDOT should continue efforts to refine and improve the design.
        
                    \noindent\textbf{Sentiment:} Positive
        
                    \noindent\textbf{Comment:} 
        
                    I want to express that we prefer alternatives 3 or 4 for Segment 1 of the North Freeway project. The construction will intensify the traffic.
        
                    \noindent\textbf{Sentiment:} Neutral
        
                    - - -
                    };
        \end{tikzpicture}
        \caption{An illustration of the 3-shot prompt along with associated sentiment labels provided to a model prompt.}
        \label{figure:ZCoT-with-answers2}
        \end{minipage}
\end{figure}

\section{Contamination Check}
Although there is no accurate approach to checking for data contamination in LLMs, we conduct a data contamination check based on the Data Contamination Quiz \cite{golchin2023data}. The contamination check is performed on a total of 30 instances randomly selected from each dataset using three LLM models. During the contamination check, we also considered both exact match and near-exact match as indicators of contamination.

To evaluate the contamination status of a dataset, we assume that data contamination has occurred if 50\% or more of all instances are contaminated. Conversely, we consider a dataset to be free of contamination if less than 50\% of the instances are identified as contaminated. Finally, a dataset is classified as contaminated if two or three models indicate contamination. Table \ref{cont} shows a summary of the performed contamination checks, with FB and NS being contaminated and SC and PH free from contamination.

% \begin{table}[h]
\begin{table}[]
\centering
\resizebox{\columnwidth}{!}{%
\begin{tabular}{ccccc}
\Xhline{0.44mm}
\textbf{Dataset} & \textbf{GPT-4o} & \begin{tabular}[c]{@{}c@{}}\textbf{Gemini}\\ \textbf{1.5 Pro}\end{tabular} & \begin{tabular}[c]{@{}c@{}}\textbf{Claude} \\ \textbf{3.5 Sonnet}\end{tabular} & \begin{tabular}[c]{@{}c@{}}\textbf{Contamination} \\ \textbf{Status}\end{tabular}  \\ \hline
FB      & × & \checkmark  & \checkmark  & \checkmark  \\
NS      & \checkmark & \checkmark  & \checkmark  & \checkmark  \\
SC      & × & ×  & ×  & ×  \\
PH      & × & ×  & ×  & ×  \\
\Xhline{0.46mm}

\end{tabular}
}
\caption{Contamination status based on a 50\% threshold. A dataset is considered contamination-free if less than 50\% of instances are identified as contaminated (×). Conversely, a dataset is marked as contaminated if 50\% or more of the instances show signs of contamination (\checkmark).}
\label{cont}
\end{table}

\section{Complexity Ranking}
To analyze and compare the complexity of the datasets, we evaluated them based on dynamic sentiment shifts and the linguistic complexity of instances. This evaluation employed expert ranking through the Borda count method. Table \ref{Borda Count Method} illustrates the complexity rankings of four datasets as assessed by three experts using this method~\cite{lansdowne1996applying}. Each expert assigned rankings according to their evaluations, and the overall ranking for each dataset was determined by summing the points assigned by all experts.

\begin{table}[h]
\centering
\resizebox{\columnwidth}{!}{%
\begin{tabular}{ccccccc}
\toprule
\textbf{Dataset} & \textbf{Expert 1} & \textbf{Expert 2} & \textbf{Expert 3} & \textbf{Total Points} & \textbf{Rank} \\ 
\midrule
\textbf{FB} & 0 & 0 & 1 & 1 & 1\\
\textbf{NS} & 2 & 3 & 2 & 7 & 3\\
\textbf{SC} & 3 & 2 & 3 & 8 & 4\\
\textbf{PH} & 1 & 1 & 0 & 2 & 2\\ 
\bottomrule
\end{tabular}%
}
\caption{Complexity ranking of datasets based on the Borda Count method, showing individual scores from each expert, total points for each dataset, and final ranks based on language complexity.}
\label{Borda Count Method}
\end{table}

\end{document}